\documentclass[conference]{IEEEtran}
\IEEEoverridecommandlockouts

\usepackage{amsmath,amssymb,amsfonts}
\usepackage{mathtools}
\usepackage{algorithmic}
\usepackage{graphicx}
\usepackage{textcomp}
\usepackage{xcolor}
\definecolor{SkyBlue}{HTML}{007FA8}
\usepackage{booktabs}
\usepackage{multirow}
\usepackage{bm}
\usepackage{pifont}
\mathtoolsset{showonlyrefs}  

\usepackage{silence}
\usepackage[font=normalsize]{caption}
\usepackage{url}
\usepackage[colorlinks=true,linkcolor=blue,citecolor=green,urlcolor=blue,bookmarks=true]{hyperref}

\newcommand{\cmark}{\ding{51}}

\newcommand{\RR}{\mathbb{R}}
\newcommand{\calN}{\mathcal{N}}

\newcommand{\bX}{\mathbf{X}}
\newcommand{\bE}{\mathbf{E}}
\newcommand{\bA}{\mathbf{A}}
\newcommand{\bM}{\mathbf{M}}
\newcommand{\bz}{\mathbf{z}}
\newcommand{\bmu}{\bm{\mu}}
\newcommand{\bsig}{\bm{\sigma}}
\newcommand{\beps}{\bm{\varepsilon}}
\newcommand{\bctx}{\mathbf{c}_{\mathrm{ctx}}}

\newcommand{\DUGO}{\textsc{DUGO}}
\newcommand{\DSE}{\textsc{DSE}}
\newcommand{\GMB}{\textsc{GMB}}
\newcommand{\OAT}{\textsc{OAT}}
\newcommand{\HAT}{\textsc{HAT}}
\newcommand{\ONMS}{\textsc{ONMS}}

\newcommand{\Lcls}{\mathcal{L}_{\mathrm{cls}}}
\newcommand{\Lreg}{\mathcal{L}_{\mathrm{reg}}}
\newcommand{\LDIoU}{\mathcal{L}_{\mathrm{DIoU}}}
\newcommand{\LKL}{\mathcal{L}_{\mathrm{KL}}}
\newcommand{\Ltot}{\mathcal{L}_{\mathrm{total}}}

\def\BibTeX{{\rm B\kern-.05em{\sc i\kern-.025em b}\kern-.08em
    T\kern-.1667em\lower.7ex\hbox{E}\kern-.125emX}}

\begin{document}


\title{DUGO: Dual-Scale Gaussian Memory for Online Temporal Action Localization in Egocentric Procedural Videos}

\author{
\IEEEauthorblockN{
Mohammed Aman Bhuiyan\textsuperscript{1},
Md. Sazzad Hossain Adib\textsuperscript{1},
Samiul Basir Bhuiyan\textsuperscript{1}, \\
Mir Faiyaz Hossain\textsuperscript{1},
Syed Mohammed Shamsul Islam\textsuperscript{2,3,4},
Md Zakir Hossain\textsuperscript{5},
Shafin Rahman\textsuperscript{1}
}

\IEEEauthorblockA{\textsuperscript{1}Department of Electrical and Computer Engineering, North South University, Dhaka, Bangladesh}

\IEEEauthorblockA{\textsuperscript{2}School of Science, Edith Cowan University, Australia}
\IEEEauthorblockA{\textsuperscript{3} School of Engineering, Computing and Mathematics, The University of Western Australia, Australia}
\IEEEauthorblockA{\textsuperscript{4} Curtin Medical School, Curtin University, Australia}

\IEEEauthorblockA{\textsuperscript{5}School of Systems and Computing, University of New South Wales, Australia}

\{mohammed.aman, sazzad.adib, samiul.bhuiyan, mir.hossain01, shafin.rahman\}@northsouth.edu \\ syed.islam@ecu.edu.au, zakir.hossain@unsw.edu.au 
}

\maketitle

\begin{abstract}
Online Temporal Action Localization (OnTAL) aims to detect and localize action instances from untrimmed streaming videos in real time. Unlike offline approaches, OnTAL models have no access to future frames and cannot go back and refine earlier predictions, which makes the task considerably harder given the strict real-time constraints and narrow temporal window available. Most existing OnTAL methods lean on either brief anchor-based context or a single-scale memory of past frames, which limits how well they can capture actions unfolding across multiple time scales or requiring long-range reasoning in Procedural Videos. To address these gaps, we present DUGO (Dual-Scale Encoding and Gaussian Memory Online), a new OnTAL framework that jointly models local motion dynamics and long-range temporal dependencies. At its core, DUGO uses a Dual-Scale Temporal Encoder, pairing depth-wise local representations with global self-attention so it can pick up both fine-grained temporal patterns and broader contextual cues. We also introduce a Gaussian Memory Bank, which represents action memory as stochastic, input-conditioned Gaussian prototypes, allowing for richer temporal memory through gated cross-attention applied at both the encoder and anchor-query stages. Extensive experiments on procedural egocentric datasets (EGTEA, EPIC-Kitchens-100) demonstrate that DUGO achieves state-of-the-art performance while also providing substantial improvements on non-egocentric benchmarks such as Muses. The project page is available at: \href{https://aman554-eq.github.io/DUGO}{\textcolor{SkyBlue}{https://aman554-eq.github.io/DUGO}}.
\end{abstract}


\section{Introduction}
\label{sec:intro}
The proliferation of video data across surveillance infrastructures, live-broadcast platforms, and safety-critical autonomous systems has intensified the demand for methods capable of semantically parsing untrimmed video streams in real time, and as a result, the demand for video understanding is growing rapidly. In particular, dealing with untrimmed videos is key to practical video understanding since most real-world videos are not pre-trimmed for individual events. Temporal Action Localization (TAL)~\cite{shou2016temporal,gao2017turn,lin2018bsn,lin2019bmn} addresses this challenge by detecting each action instance in an input untrimmed video through the prediction of its start time, end time, and action class. Recent advances in TAL have leveraged graph-relational reasoning~\cite{xu2020gtad,zeng2019graph,zhao2021vsgn}, temporal multi-scale feature pyramids~\cite{liu2020progressive}, global context enrichment~\cite{zhu2021enriching}, and long-memory Transformers~\cite{xu2021lstr} to achieve impressive boundary precision. More recently, vision-language models have been used for open-vocabulary TAL, but they often exhibit a modality bias toward textual rather than visual cues \cite{li2026towards}. Yet virtually all such methods presuppose that the full video is observable before inference, an assumption irreconcilable with real-time streaming constraints.

To bridge this gap, the field has explored several online paradigms. Online Action Detection (OAD)~\cite{degeest2016online} casts recognition as causal frame-wise classification, with subsequent extensions employing reinforced learning~\cite{gao2017red}, future-frame anticipation~\cite{kim2021temporally}, and Transformer-based history encoders~\cite{xu2021lstr,chen2022gatehub}, and, more recently, hierarchical memory-compression mechanisms that adaptively retain critical long-range context under a bounded storage budget~\cite{yan2026backtrace}. Online Detection of Action Start (ODAS)~\cite{shou2018online,gao2019startnet} further prioritizes temporal responsiveness by targeting action onset. That said, both OAD and ODAS operate at the \emph{frame level}, meaning they fall short of the instance-level semantic grounding that downstream tasks like activity summarization, event-driven retrieval, and structured scene understanding actually need. CAG-QIL~\cite{kang2021cagqil} formally defined Online Temporal Action Localization (OnTAL), an instance-level online task subject to two stringent constraints: \emph{(i)} only past and current frames are accessible, and \emph{(ii)} committed proposals are immutable, precluding retrospective refinement via Non-Maximum Suppression (NMS). Their method groups per-frame OAD predictions into instances via imitation-learning-trained Markov Decision Processes, but inherits three structural failure modes: ticks, fragmentation, and merging.

\begin{figure*}[t]
  \centering
  \includegraphics[width=\textwidth]{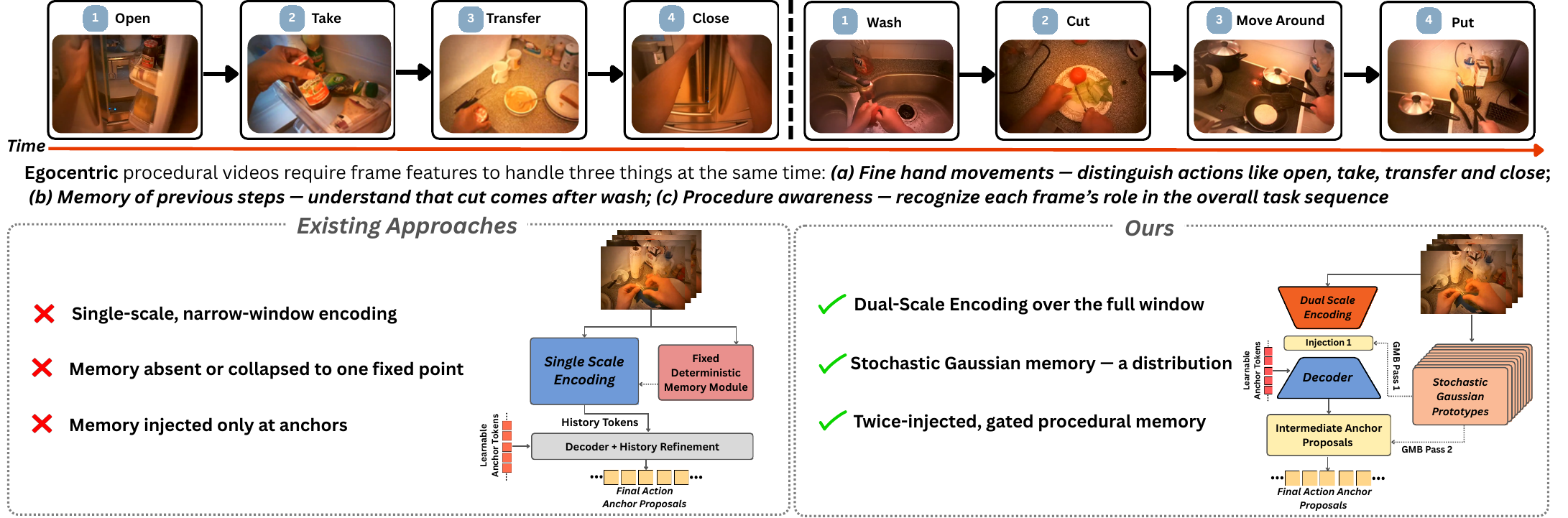}
  \caption{Motivation. Egocentric procedural videos require fine-grained motion discrimination, memory of prior steps, and procedure-aware framing. Prior OnTAL methods rely on single-scale encoding with absent or one-shot deterministic memory; DUGO introduces dual-scale encoding and a twice-injected stochastic Gaussian memory that preserves multi-modal procedural ambiguity.
  }
  \label{fig:overview}
\end{figure*}

Online Anchor Transformer (\OAT{})~\cite{kim2022oat} addressed these limitations by proposing the sliding-window-based Anchor Transformer framework, which abandons OAD-based grouping in favor of a sliding-window anchor mechanism. A Transformer encoder converts queued features into temporally contextualized representations, while a Transformer decoder synthesizes $M$ anchor features using learnable anchor queries. Each anchor is refined by a dual-head module that predicts the action class, end offset, and length ratio. \OAT{}'s~\cite{kim2022oat} receptive field, however, is confined to the most recent sliding window, discarding all prior stream evidence. History-Augmented Anchor Transformer (\HAT{})~\cite{reza2024hat} identified this limitation as particularly severe in procedural and egocentric scenarios, where actions are causally conditioned on distant historical context. \HAT{}~\cite{reza2024hat} augments \OAT{}~\cite{kim2022oat} with a Future-Supervised History Module, which compresses long-term history through cross-attention over learnable tokens, calibrates the compressor using an action anticipation head, and refines the compressed history against short-term window encoding.

Despite these advances, critical limitations persist (see Figure~\ref{fig:overview}). \OAT{}~\cite{kim2022oat} remains restricted to a fixed frame sliding window and entirely lacks explicit procedural memory, preventing temporal disambiguation from prior action context. \HAT{}~\cite{reza2024hat} offers a partial fix by compressing history deterministically, but in doing so it flattens the naturally multi-modal space of procedural action trajectories into a single latent representation, losing ambiguity and any sense of alternative procedural paths along the way. On top of that, its single-granularity encoding of history cannot hold onto fine-grained motion cues and long-range semantic dependencies at the same time. To address these shortcomings, we propose \DUGO{}, an OnTAL framework that combines dual-scale temporal encoding with stochastic procedural memory modeling, jointly capturing fine-grained motion dynamics and long-range action dependencies while still preserving the multi-modal ambiguity present in the action history. In addition, inspired by DIoU loss \cite{zheng2020DIoU}, we incorporated a 1D DIoU loss to reduce localization error through improved temporal boundary refinement and center alignment.
Our principal contributions are as follows:
\begin{itemize}
  \item We propose \DUGO{}, a novel online temporal action localization architecture that jointly integrates dual-scale temporal encoding and stochastic procedural memory modeling. By combining local convolutional dynamics with global self-attention and augmenting them with adaptive Gaussian-based memory representations, the framework captures both fine-grained motion patterns and long-range procedural dependencies while preserving past context for procedural tasks.
  \item We incorporate 1D DIoU loss into the OnTAL regression head, reducing localization error through improved regression boundary refinement by jointly optimizing temporal overlap and center-point alignment.
  \item Extensive experiments on EGTEA and EPIC-Kitchens-100 establish a new state-of-the-art under the online constraint. DUGO also demonstrates consistent improvements on non-egocentric benchmarks, with particularly strong gains on procedural egocentric datasets where multi-scale temporal encoding and stochastic memory modeling prove most effective.
\end{itemize}

\section{Related Work}

\noindent\textbf{Online Video Understanding:}
Online video understanding requires models to process streaming video sequentially
without access to future context. The foundational task is Online Action Detection
(OAD)~\cite{degeest2016online}, which classifies the
action occurring in each incoming frame in real time. Subsequent works improved OAD performance using several approaches, including reinforced learning~\cite{gao2017red}, future-frame anticipation~\cite{kim2021temporally}, transformer-based context
encoders~\cite{chen2022gatehub}, and long short-term memory architectures~\cite{xu2021lstr}
that efficiently compress extended history queues into compact temporal
representations. A related but distinct task is Online Detection of Action Start
(ODAS)~\cite{shou2018online}, which aims to identify the exact frame at which an action begins. Although these approaches provide strong frame-level temporal representations for online settings, their frame-wise predictions do not provide the instance-level temporal grounding needed for downstream procedural applications. Moreover, most of them model temporal context at a single scale, which can make it difficult to capture actions whose distinguishing cues occur across different temporal durations. DUGO addresses these limitations through instance-level reasoning and a
dual-scale temporal encoder that sequentially combines a depthwise convolutional
local branch with a global self-attention branch, jointly capturing fine-grained
motion dynamics and long-range temporal dependencies within an  encoder stage.

\noindent\textbf{Online Temporal Action Localization:}
Temporal action localization (TAL) identifies action instances and their boundaries
in untrimmed video. Early sliding window approaches~\cite{shou2016temporal,gao2017turn}
process fixed segments independently; subsequent works expand contextual coverage
through boundary-sensitive proposal generation~\cite{lin2018bsn,lin2019bmn},
graph-based inter-proposal reasoning~\cite{zeng2019graph,xu2020gtad}, and
with multi-scale feature pyramids~\cite{liu2020progressive}. This clip-wise processing cost remains a bottleneck even for language-grounded moment retrieval in long videos: DeCafNet \cite{lu2025decafnet} addresses it offline via a lightweight delegate encoder that filters clips before a full-scale expert encoder processes only the salient ones.  All offline TAL methods, however, assume
full-video access, making them unsuitable for streaming deployments. Online Temporal
Action localization (OnTAL), first formalized by CAG-QIL \cite{kang2021cagqil}, requires the model to generate complete proposals from
a streaming prefix without modifying previously committed decisions. CAG-QIL groups per-frame OAD predictions into instances via a Markov Decision Process.
\OAT{}~\cite{kim2022oat} identified fragmentation and merging artefacts in
grouping-based approaches and proposed an anchor-based Transformer that applies
instance-level supervision within a sliding window, replacing offline NMS with the
streaming-compatible \ONMS{}. \HAT{}~\cite{reza2024hat} subsequently augmented \OAT{}~\cite{kim2022oat}
with a deterministic history module injected once into the anchor features. MATR~\cite{song2024online}
further extends OnTAL with a memory queue that selectively stores past segment features and separately
retrieves historical context to localize action starts, but its memory-centric boundary localization is not designed for the multi-step, ambiguous trajectories of egocentric procedural tasks. OpenHOUSE \cite{kang2025openhouse} pushes this streaming setting further still by pairing class-agnostic On-TAL with free-form, LLM-grounded description generation, enabling hierarchical event understanding beyond a fixed action vocabulary.
Existing OnTAL methods are either confined to a fixed short-term window with no procedural memory, or rely on a
single-stage deterministic history injection that discards multi-modal action trajectory ambiguity.
In contrast, \DUGO{} extends the anchor-transformer paradigm with dual-scale encoding and
two-stage stochastic history injection, preserving richer temporal context throughout the
encoder-decoder pipeline.

\noindent\textbf{Memory for Online Action Understanding:}
Integrating long-term historical context into online video models is a well-established
strategy for improving temporal reasoning. Within OAD, LSTR~\cite{xu2021lstr}
compresses an extended history queue into a compact representation via a two-stage
transformer decoder, enabling efficient long-range context integration. GateHUB~\cite{chen2022gatehub}
augments one-stage history compression with a gating module that suppresses irrelevant
background frames and calibrates attention over past observations. \HAT{}~\cite{reza2024hat}
brought structured history modeling into OnTAL, introducing a Future-Supervised History
Module that employs a learnable token-based transformer decoder with future-driven alignment.
MATR~\cite{song2024online} similarly maintains a selective memory queue of past segment features,
but its current-segment/end and memory/start decoding scheme is primarily designed for action-boundary
localization rather than the fine-grained, multi-modal dependencies characteristic of procedural egocentric activities.
Evaluations on procedural egocentric benchmarks~\cite{li2018egtea,damen2022epickitchens} confirm
that historical context yields substantial gains for structured, sequential action scenarios.
Despite these advances, all prior history modules rely on deterministic compression, collapsing
the exponentially large space of procedural action trajectories into a single fixed-length vector
and discarding distributional ambiguity essential for procedural reasoning. \DUGO{}
replaces deterministic compression with a stochastic mixture of input-conditioned Gaussian
prototypes injected at two pipeline stages via per-head gated cross-attention, preserving
multi-modal procedural ambiguity that all prior history approaches discard.

\begin{figure*}[t]
  \centering
  \includegraphics[width=\textwidth]{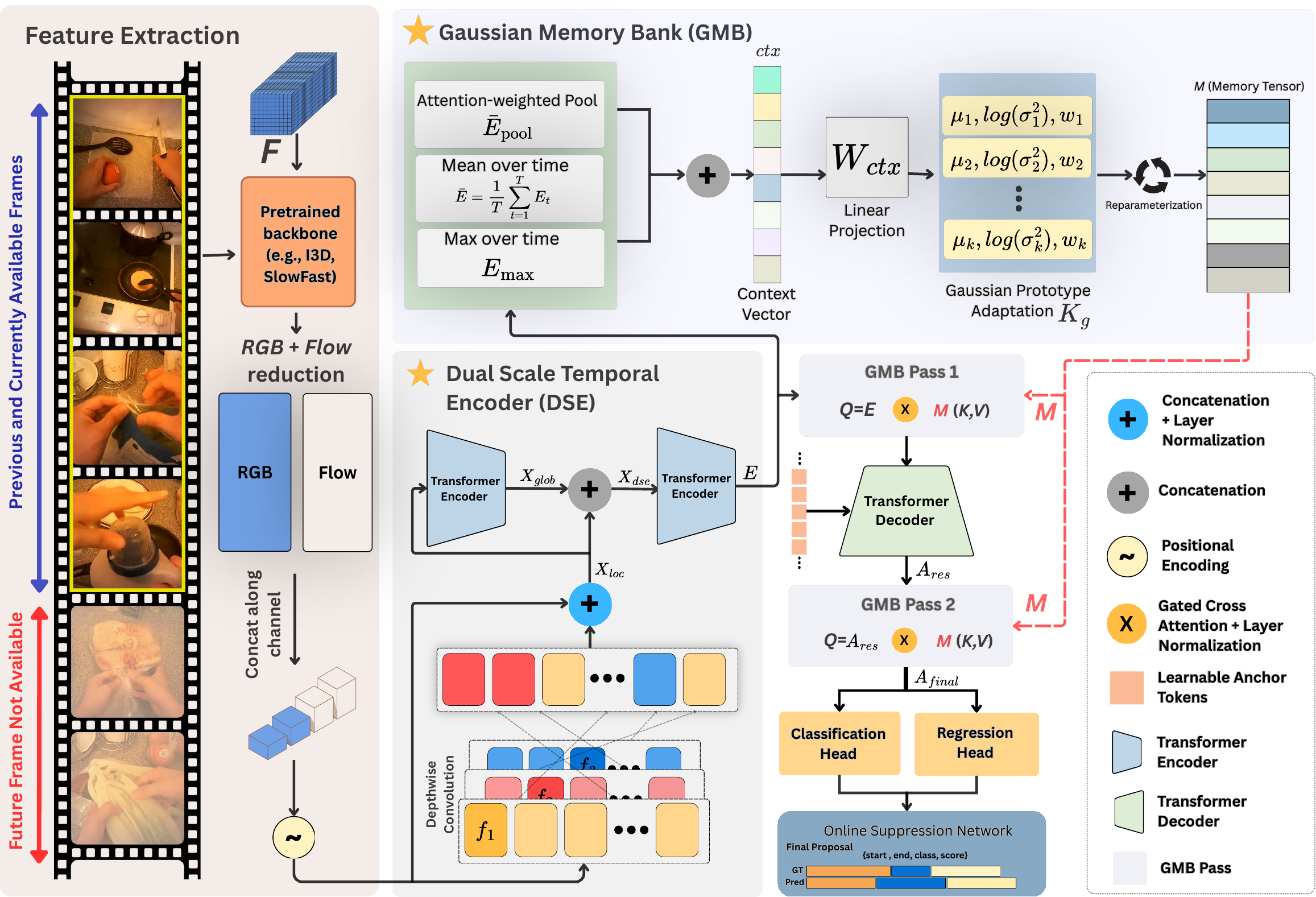}
  \caption{\textbf{\DUGO{} Architecture.}
    \DSE{} replaces the single-granularity encoder with dual branches consisting of
    depthwise temporal convolutions for local modeling and self-attention for global
    temporal reasoning, followed by fusion through a learned projection.
    \GMB{} replaces deterministic history compression with a stochastic mixture of
    Gaussian prototypes injected at two stages: encoder memory refinement (Pass~1)
    and decoded anchor query enhancement (Pass~2) through per-head gated cross-attention.
    }
  \label{fig:architecture}
\end{figure*}

\section{Methodology}
\label{sec:method}

\subsection{Problem Formulation and Feature Representation}
\label{sec:problem_feat}

\noindent\textbf{Problem Definition:}
Let $V = \{v_i\}_{i=1}^{T}$ denote an untrimmed egocentric video and
$\Psi = \{(s_k, e_k, c_k)\}_{k=1}^{K}$ be the $K$ action instances with start time
$s_k$, end time $e_k$, and class label $c_k$.
Under the OnTAL constraint, at timestamp $t$ only the partial video $V_{1:t}$ is
observable. The goal is to generate action proposals immediately upon action-end
detection, sequentially recovering $\Psi$ without modifying previously emitted proposals.

\noindent\textbf{Feature Extraction and Positional Encoding:}
At timestamp $t$, a sliding window of $L_q$ frames yields the feature queue
$\mathbf{F} \in \RR^{L_q \times 2D_f}$, where each frame concatenates an RGB descriptor
and an optical-flow descriptor of dimension $D_f$ each.
The two streams are independently projected from $D_f$ to $D/2$ via learned linear layers
$\mathbf{W}_{\mathrm{rgb}}$ and $\mathbf{W}_{\mathrm{flow}}$, then concatenated to form
$\bX \in \RR^{L_q \times D}$.
Standard sinusoidal positional encoding is added channel-pair-wise,
$\mathrm{PE}_{i,2j} = \sin(i/10000^{2j/D})$ and $\mathrm{PE}_{i,2j+1} = \cos(i/10000^{2j/D})$
for temporal position $i$ and channel pair $j$, yielding $\bX_{\mathrm{pe}} \in \RR^{L_q \times D}$.

\noindent\textbf{Overall Architecture:}
Figure~\ref{fig:architecture} illustrates the full \DUGO{} pipeline.
At each timestamp $t$, the positionally encoded features $\bX_{\mathrm{pe}}$ are processed
by two principled modules.
The Dual-Scale Temporal Encoder (\DSE{}) captures both fine-grained local motion
dynamics and long-range action dependencies within the $L_q$-frame observation window,
producing the encoder memory $\bE \in \RR^{L_q \times D}$.
The Gaussian Memory Bank (\GMB{}) models procedural action history as a stochastic
mixture of $K_g$ learnable Gaussian prototypes conditioned on $\bE$, injecting history
at two stages via per-head gated cross-attention: first into the encoder memory and then into
the decoded anchor features.
The resulting anchor representations $\bA_{\mathrm{final}} \in \RR^{M \times D}$ feed
classification and regression heads and are post-processed by \ONMS{} to produce the final
output $\Psi$.

\subsection{Dual-Scale Temporal Encoder}
\label{sec:dse}

Standard Transformer encoders apply self-attention uniformly across all positions
in a context window, operating at a single temporal granularity.
This global-only processing dilutes sharp, transient signals such as sudden hand
velocity changes or brief contact events that define precise action boundaries in
egocentric procedural video.
Conversely, a convolution-only design captures local patterns but lacks the receptive
field to model how the current moment relates to events from the previous several frames.
\DSE{} addresses this by processing the positionally encoded input
$\bX_{\mathrm{pe}} \in \mathbb{R}^{L_q \times D}$ through a sequential two-branch
design, where $L_q$ denotes the observation window length and $D$ denotes the
feature dimension.

\noindent\textbf{Local Branch:}
The local branch applies a depthwise 1D convolution $\mathcal{F}_l(\cdot)$ with
kernel size $k$, where each of the $D$ feature channels is convolved with its own
independent filter.
In procedural egocentric video, actions such as \emph{open}, \emph{take}, and
\emph{close} are visually similar yet differ in subtle wrist rotations and grip
transitions spanning only a few frames; the depthwise design preserves these
per-channel dynamics without cross-channel mixing that could dilute them.
Padding is applied to preserve the sequence length.
A residual connection and layer normalization stabilize the output:
\begin{equation}
  \bX_{\mathrm{loc}}
    = \mathrm{LN}\!\bigl(\mathcal{F}_l(\bX_{\mathrm{pe}}) + \bX_{\mathrm{pe}}\bigr)
    \;\in \mathbb{R}^{L_q \times D},
  \label{eq:local}
\end{equation}
where $\mathrm{LN}(\cdot)$ denotes layer normalization.
The residual connection ensures that if the convolution provides no discriminative
signal, the original representation passes through unchanged, preventing degradation
at ambiguous frames.

\noindent\textbf{Global Branch:}
Procedural tasks are inherently sequential: the meaning of a pour depends on what
was cracked before it, and whether a spread is complete depends on context from earlier
in the window.
To capture these inter-frame structural dependencies, the global branch applies
$\mathcal{F}_g(\cdot)$, a Transformer encoder layer with multi-head self-attention
followed by a position-wise feed-forward network, to the locally refined
$\bX_{\mathrm{loc}}$:
\begin{equation}
  \bX_{\mathrm{glob}}
    = \mathcal{F}_g\!\bigl(\bX_{\mathrm{loc}}\bigr)
    \;\in \mathbb{R}^{L_q \times D}.
  \label{eq:global}
\end{equation}
By receiving $\bX_{\mathrm{loc}}$ rather than the raw input, $\mathcal{F}_g$
specializes in modeling structural dependencies over locally refined features,
rather than operating directly on raw temporal signals.
Each of the $L_q$ frames attends to all other positions, giving every representation
a window-wide temporal context.

\noindent\textbf{Fusion and Encoding:}
The two representations are concatenated along the feature dimension and projected
back to $D$ through a learned fusion block $\mathcal{F}_f(\cdot)$:
\begin{equation}
  \bX_{\mathrm{dse}}
    = \mathcal{F}_f\!\bigl([\bX_{\mathrm{loc}};\,\bX_{\mathrm{glob}}]\bigr)
    \;\in \mathbb{R}^{L_q \times D},
  \label{eq:dse}
\end{equation}
where $[\,\cdot\,;\,\cdot\,]$ denotes concatenation along the feature axis and
$\mathbf{W}_f \in \mathbb{R}^{2D \times D}$ is the learned projection within
$\mathcal{F}_f(\cdot)$.
$\mathbf{W}_f$ learns a per-dimension mixing of local and global features, allowing
different output dimensions to weight the two branches independently.
$\bX_{\mathrm{dse}}$ is then passed through stacked Transformer encoder blocks to
produce the encoder memory $\bE \in \mathbb{R}^{L_q \times D}$, which serves as
the context for both the anchor decoder and the Gaussian Memory Bank.

\subsection{Gaussian Latent Memory for Online Procedural Modeling}
\label{sec:GMB}

The \GMB{} (Gaussian Memory Bank) models procedural action history as a stochastic mixture of $K_g$ learned
Gaussian prototypes, each representing a distinct action archetype.
Rather than compressing raw frame history, \GMB{} adapts these prototypes
to the current encoder context, yielding a compact and regularized history representation.

\noindent\textbf{Temporal context summarization:}
Given encoder output $\bE \in \RR^{L_q \times D}$, a context vector
$\bctx \in \RR^{D}$ is formed by fusing three complementary pooling signals: an
attention-weighted pool computed using the temporal mean $\bar{\bE}$ as the query over
$\bE$, the temporal mean $\bar{\bE}$ itself, and the temporal maximum $\max(\bE)$.
These capture ordered temporal structure, global statistics, and peak activations
respectively, and are concatenated and projected by a learned matrix
$\mathbf{W}_{\mathrm{ctx}} \in \RR^{3D \times D}$ to produce $\bctx$.

\noindent\textbf{Input-conditioned prototype adaptation:}
A global prior dictionary
$\{\bmu^{\mathrm{prior}}_k,\,\log\bsig^{2,\mathrm{prior}}_k,
\,w^{\mathrm{prior}}_k\}_{k=1}^{K_g}$
is stored as learnable parameters.
Log-scale variance is used because $\log\sigma^2$ is unconstrained over $\mathbb{R}$,
ensuring $\sigma^2 = \exp(\log\sigma^2) > 0$ at all times without requiring constrained
optimization, and yielding numerically stable gradient computation through the
reparameterization step.
Zero-initialized linear heads compute per-input corrections from $\bctx$, scaled by a
learned adaptation gate $g = \sigma(\alpha_{\mathrm{gate}})$ initialized so that $g \approx 0.12$,
keeping the prior dominant early in training:
\begin{align}
  \bmu_k        &= \bmu^{\mathrm{prior}}_k
                   + g\cdot\Delta\bmu_k(\bctx),              \notag \\
  \log\bsig^2_k &= \log\bsig^{2,\mathrm{prior}}_k
                   + g\cdot\Delta\log\bsig^2_k(\bctx),       \notag \\
  \mathbf{w}    &= \operatorname{Softmax}\!\bigl(
                   \mathbf{w}^{\mathrm{prior}}
                   + g\cdot\Delta\mathbf{w}(\bctx)\bigr),
  \label{eq:adapt}
\end{align}
where $\bmu_k,\,\log\bsig^2_k \in \RR^{D}$ are the adapted mean and log-variance for
prototype $k$, $\mathbf{w} \in \RR^{K_g}$ are the normalized mixture weights, and
$\Delta\bmu_k$, $\Delta\log\bsig^2_k$, $\Delta\mathbf{w}$ are linear projections of
$\bctx$ with zero-initialized weights.
During training, samples $\bz_k$ are drawn via the reparameterization trick,
$\bz_k = \bmu_k + \beps_k \odot \exp(0.5\cdot\log\bsig^2_k)$ with
$\beps_k \sim \calN(\mathbf{0},\mathbf{I})$; at inference $\bz_k = \bmu_k$ directly.

\noindent\textbf{Gaussian memory tensor:}
Each prototype contributes one weighted sample, and the $K_g$ samples are stacked to
form the memory bank:
\begin{align}
  \bM_k &= w_k \cdot \bz_k \;\in \RR^{D},
           \quad k = 1,\ldots,K_g,                           \notag \\
  \bM   &= [\bM_1;\;\ldots;\;\bM_{K_g}] \;\in \RR^{K_g \times D},
  \label{eq:bank}
\end{align}
where $w_k \in \mathbb{R}$ is the mixture weight for prototype $k$ drawn from
$\mathbf{w}$.
Stacking rather than summing preserves each prototype as an individually addressable
memory slot during subsequent cross-attention.

\noindent\textbf{GMB cross-attention with per-head gating:}
Let $\mathbf{x} \in \RR^{S \times D}$ denote the input sequence to be augmented,
where $S = L_q$ when applied to encoder memory and $S = M$ when applied to anchor
queries.
Given memory bank $\bM \in \RR^{K_g \times D}$, the \GMB{} cross-attention module
computes multi-head cross-attention $\operatorname{MHCA}(\mathbf{x},\bM,\bM)$ with
$\mathbf{x}$ as queries and $\bM$ as keys and values, then applies a per-head
cross-attention gate before the residual connection:
\begin{equation}
  \operatorname{GMB}(\mathbf{x},\bM) =
    \operatorname{LN}\!\bigl(
      \mathbf{x} + \mathbf{G} \odot \operatorname{MHCA}(\mathbf{x},\bM,\bM)
    \bigr)
    \;\in \RR^{S \times D}.
  \label{eq:GMB}
\end{equation}
The gate vector $\mathbf{G} \in [0,1]^D$ is the cross-attention gate, constructed head-wise: the $d_h = D/H$
entries for attention head $h$ are each set to $\sigma(g_h)$, where $g_h$ is a scalar
parameter initialized to zero, giving $\sigma(0) = 0.5$ at the start of training.
Each head independently learns to increase or decrease its history contribution,
allowing the model to selectively rely on procedural memory on a per-head basis.

\noindent\textbf{Two-stage GMB injection:}
\GMB{} is applied at two distinct points in the forward pass, sharing the same
prototype dictionary and adaptation heads at both call sites, incurring no additional
parameters.
In \textbf{Pass~1}, the encoder memory $\bE$ is augmented with Gaussian history
context before decoding:
\begin{equation}
  \bE_{\mathrm{GMB}} = \operatorname{GMB}\!\bigl(\bE,\;\bM(\bE)\bigr)
    \;\in \RR^{L_q \times D},
  \label{eq:pass1}
\end{equation}
where $\bM(\bE)$ denotes the bank conditioned on $\bE$ itself.
The anchor decoder then cross-attends to $\bE_{\mathrm{GMB}}$, decoding $M$ learnable
anchor query vectors $\mathbf{Q}_{\mathrm{anc}} \in \RR^{M \times D}$ to obtain:
\begin{equation}
  \bA_{\mathrm{res}} = \operatorname{LN}\!\bigl(
    \operatorname{TransDec}(\mathbf{Q}_{\mathrm{anc}},\,\bE_{\mathrm{GMB}})
    + \mathbf{Q}_{\mathrm{anc}}\bigr)
  \;\in \RR^{M \times D}.
  \label{eq:ares}
\end{equation}
In \textbf{Pass~2}, \GMB{} is applied again on $\bA_{\mathrm{res}}$, with the bank
always conditioned on the full encoder output $\bE$ so that prototype adaptation
observes all $L_q$ temporal positions:
\begin{equation}
  \bA_{\mathrm{final}} = \operatorname{GMB}\!\bigl(\bA_{\mathrm{res}},\;\bM(\bE)\bigr)
    \;\in \RR^{M \times D}.
  \label{eq:pass2}
\end{equation}
The two injections are complementary: Pass~1 enriches the encoder memory the decoder
attends to, while Pass~2 injects history directly into the final anchor features that
feed the prediction heads.

\subsection{Prediction and Optimization}
\label{sec:pred}

\noindent\textbf{Prediction module and post-processing:}
Classification and regression heads, each a two-layer feed-forward network, operate on
$\bA_{\mathrm{final}} \in \RR^{M \times D}$ to produce class logits
$\mathbf{z}_c \in \RR^{M \times (C{+}1)}$ spanning $C{+}1$ classes including background,
and regression offsets $\mathbf{z}_r \in \RR^{M \times 2}$.
At inference, the predicted end and start times for anchor $i$ are decoded as
\begin{equation}
    e_i = a_{o,i} + a_{l,i}\cdot z_{r,i}^{(0)},
    \qquad
    s_i = e_i - a_{l,i}\cdot\exp\!\bigl(z_{r,i}^{(1)}\bigr),
    \label{eq:decode}
\end{equation}
where $a_{o,i}$ is the anchor origin (nominal end position), $a_{l,i}$ is the anchor length,
$z_{r,i}^{(0)}$ is the predicted end-point offset relative to anchor length,
and $z_{r,i}^{(1)}$ is the predicted log length ratio.

At each timestamp $t$, the $M$ decoded proposals are filtered by an Online Suppression
Network (\ONMS{}), a lightweight streaming alternative to offline NMS\@. The \ONMS{} takes as
input a per-class confidence matrix $\mathbf{P}_{\mathrm{osc}} \in \RR^{L_q \times C}$
accumulated over the last $L_q$ frames and produces a per-class suppression score
$\mathbf{s} \in [0,1]^C$ via a two-layer MLP applied independently to each class
trajectory. Proposals below the suppression threshold $\theta_s{=}0.1$ are discarded.
The \ONMS{} is trained separately after the main model is frozen, using binary proposal
labels derived from the main model's predictions on the training set.

\noindent \textbf{Focal classification loss:}
The $M$ anchor predictions are matched to ground-truth intervals via IoU threshold
$\theta_m{=}0.5$, yielding a foreground set $\mathrm{FG}$ and a background set. To address
the heavy foreground-to-background class imbalance, we apply focal cross-entropy with
modulating factor $\gamma{=}2$:
\begin{equation}
  \Lcls = \sum_{i=1}^{M}\sum_{j=0}^{C}
    -\tilde{y}_{c,i}^{\,j}\,\bigl(1 - p_{c,i}^{\,j}\bigr)^{2}
    \log p_{c,i}^{\,j},
  \label{eq:cls_loss}
\end{equation}
where $p_{c,i}^j = \operatorname{softmax}(\mathbf{z}_{c,i})_j$ is the predicted probability
for anchor $i$ and class $j$, and
$\tilde{y}_{c,i}^j = y_{c,i}^j / \sum_{j'} y_{c,i}^{j'}$
is the multi-hot ground-truth label normalized by the number of positive class matches for
anchor $i$, ensuring per-anchor supervision sums to one.

\noindent \textbf{Regression losses:}
For foreground anchors, the L1 regression loss $\Lreg$ supervises two offsets: the
normalized end-point offset $z_{r,i}^{(0)}$ against the target
$(g_{e,i} - a_{o,i})/a_{l,i}$, and the log length ratio $z_{r,i}^{(1)}$ against
$\log(g_{l,i}/a_{l,i})$, where $g_{e,i}$ and $g_{l,i}$ are the ground-truth action end
time and length. Normalizing by anchor length $a_{l,i}$ makes the loss anchor-scale
agnostic. L1 loss treats the two offsets independently without penalizing geometric misalignment
between predicted and target segments. We therefore incorporate a 1D Distance-IoU loss that
directly maximizes temporal overlap while penalizing center misalignment. For foreground
anchor $i$, the predicted and ground-truth segments are decoded anchor-relative from the
regression logits as $(\hat{s}_i, \hat{e}_i)$ and $(s^{gt}_i, e^{gt}_i)$. Temporal IoU
is computed from the overlap and union of these decoded segments. Let $\hat{m}_i$ and
$m^{gt}_i$ denote the predicted and ground-truth segment centers, and $c_i$ the length of
the smallest enclosing interval. The DIoU penalty combines IoU with a normalized squared
center distance:
\begin{equation}
  \resizebox{\columnwidth}{!}{$\displaystyle
  \mathrm{DIoU}_i = \mathrm{IoU}_i - \frac{(\hat{m}_i - m^{gt}_i)^2}{c_i^2},
  \LDIoU = \frac{1}{|\mathrm{FG}|}\sum_{i \in \mathrm{FG}}(1 - \mathrm{DIoU}_i).
  $}
  \label{eq:DIoU_loss}
\end{equation}

\noindent \textbf{KL regularization and total loss:}
To prevent the global Gaussian prior from collapsing and ensure diverse prototype
representations, the prior parameters $\bmu^{\mathrm{prior}}$ and $\log\bsig^{2,\mathrm{prior}}$
are regularized with a KL divergence against the standard normal,
$\LKL = -(1/2K_g D)\sum_{k,d}(1 + \log\sigma^{2,\mathrm{prior}}_{k,d} - (\mu^{\mathrm{prior}}_{k,d})^2 - \sigma^{2,\mathrm{prior}}_{k,d})$,
which encourages globally diverse archetypes without over-constraining input-specific
adaptation. The four losses are combined with fixed scalar weights:
\begin{equation}
  \Ltot = \alpha\,\Lcls + \beta\,\Lreg + w_d\,\LDIoU + w_{kl}\,\LKL
  \label{eq:total}
\end{equation}
with $\alpha{=}\beta{=}1$, $w_d{=}1.0$, and $w_{kl}{=}10^{-4}$.
The small KL weight prevents the regularization term from dominating the task-specific
objectives while still maintaining prototype diversity throughout training.

\section{Experiments}

\subsection{Datasets and Implementation Details}
\label{sec:datasets}
We evaluate our method on two procedural egocentric (PREGO) datasets, EGTEA and EPIC-Kitchens-100, as well as the standard non-PREGO temporal action localization benchmark MUSES.

\noindent\textbf{EGTEA~\cite{li2018egtea}:}
EGTEA is a procedural egocentric action dataset containing approximately 28 hours of first-person video collected from 86 recording sessions involving 32 subjects performing kitchen activities. The dataset includes 22 action classes, with an average of 180 action instances and 15.6 action categories per video. Due to its sequential procedural structure and egocentric viewpoint, EGTEA provides a challenging benchmark for evaluating history-aware Online Temporal Action Localization (OnTAL) methods. Following prior work, we report mAP at temporal IoU thresholds from 0.1 to 0.5 with a step size of 0.1, denoted as mAP@[0.1:0.1:0.5], together with the averaged mAP.








\begin{table*}[t]
\centering
\footnotesize
\setlength{\tabcolsep}{4pt}
\caption{Comparison of mAP (\%) across tIoU thresholds on EGTEA and EPIC-Kitchens-100 datasets. $^{*}$ Indicates best result.}
\label{tab:egtea_epic}

\begin{tabular*}{\textwidth}{@{\extracolsep{\fill}}lcccccc|cccccc@{}}
\toprule
& \multicolumn{6}{c|}{\textbf{EGTEA~\cite{li2018egtea}}}
& \multicolumn{6}{c}{\textbf{EPIC-Kitchens-100~\cite{damen2022epickitchens}}} \\

\cmidrule(lr){2-7}\cmidrule(lr){8-13}
\textbf{Model}
& \textbf{0.1} & \textbf{0.2} & \textbf{0.3}
& \textbf{0.4} & \textbf{0.5} & \textbf{Avg.}
& \textbf{0.1} & \textbf{0.2} & \textbf{0.3}
& \textbf{0.4} & \textbf{0.5} & \textbf{Avg.} \\
\midrule

MATR~\cite{song2024online}
& 21.5 & 19.5 & 15.9 & 12.6 & 8.6 & 15.6
& 17.2 & 14.9 & 12.8 & 10.0 & 8.7 & 12.7 \\

OAT~\cite{kim2022oat}
& 24.9 & 23.1 & 20.6 & 17.1 & 12.2 & 19.6
& 17.8 & 16.3 & 14.3 & 12.8 & 10.1 & 14.2 \\

HAT~\cite{reza2024hat}
& 27.5 & 25.4 & 22.6 & 18.3 & 13.5 & 21.5
& 18.3 & 17.0 & 15.8 & 13.9 & 11.5 & 15.3 \\

\midrule

\textbf{Ours}
& \textbf{30.9} & \textbf{28.8} & \textbf{25.6}
& \textbf{21.1} & \textbf{14.9} & \textbf{24.3}$^{*}$
& \textbf{20.1} & \textbf{18.9} & \textbf{17.0}
& \textbf{15.0} & \textbf{11.7} & \textbf{16.6}$^{*}$ \\

\bottomrule
\end{tabular*}

\end{table*}

\noindent\textbf{EPIC-Kitchens-100~\cite{damen2022epickitchens}:}
EPIC-Kitchens-100 is a large-scale egocentric action recognition benchmark consisting of nearly 100 hours of video captured across 700 recording sessions in diverse kitchen environments. The dataset contains an average of 121.5 action instances and 14.5 action classes per video. Actions are annotated using verb-object pairs; following the standard OnTAL evaluation protocol~\cite{reza2024hat}, we utilize only the verb-level annotations (97 verb classes). Models are trained on the training split and evaluated on the validation split. Performance is measured using mAP@[0.1:0.1:0.5].

\noindent\textbf{MUSES~\cite{liu2021muses}:} MUSES is a large-scale non-egocentric temporal action localization benchmark comprising 3,697 videos with a total duration of approximately 716 hours. It contains 25 action classes drawn from TV shows and movie dramas, featuring diverse scenes, frequent shot cuts, and weaker procedural dependencies compared to PREGO datasets. On average, each video contains 8.5 action instances and 3.3 action classes. Following the standard evaluation protocol, we report mAP at temporal IoU thresholds ranging from 0.3 to 0.7 with a step size of 0.1, denoted as mAP@[0.3:0.1:0.7].

\noindent\textbf{Feature Encoders:}
For EGTEA, we extract video features using I3D~\cite{carreira2017i3d}, pre-trained on Kinetics. The frames are processed at the original 24 FPS with a stride of 12, producing two feature vectors per second. We use the two-stream (RGB+Flow) descriptors, resulting in an input feature dimension of 2048. For EPIC-Kitchens-100, we use SlowFast~\cite{feichtenhofer2019slowfast} features following the standard settings adopted in recent TAL studies. Specifically, we process 32-frame clips at 30 FPS with a stride of 16 frames, which corresponds to approximately one feature vector every 0.53 seconds. The resulting input feature dimension is 2304. For MUSES, we use I3D~\cite{carreira2017i3d} features at a single-stream feature dimension of 1024, consistent with prior work~\cite{reza2024hat} on the dataset.

\noindent\textbf{Network Configuration:}
All transformer modules operate with feature dimension $D{=}1024$. The \DSE{} consists of a depthwise 1D convolution branch (kernel $k{=}5$) and a Transformer encoder layer with $H{=}8$ heads, fused by a learned projection. The main encoder has $N_e{=}3$ blocks with $H{=}8$ heads,
and the anchor decoder has $N_d{=}5$ blocks with $H{=}4$ heads. The \GMB{} maintains $K_g{=}8$
Gaussian prototypes with $H{=}8$ cross-attention heads.

\noindent\textbf{Dataset-Specific Parameters:}
For EGTEA and EPIC-Kitchens-100, we set the observation window $L_q{=}64$ frames with $M{=}6$
anchor sizes of $\{2, 4, 6, 8, 12, 16\}$. For MUSES, we use a longer observation window
$L_q{=}150$ with $M{=}7$ anchor sizes of $\{4, 9, 18, 37, 75, 112, 150\}$, reflecting the
longer action durations in TV/movie content compared to procedural egocentric kitchen activity.

\noindent\textbf{Training Settings:}
The main network is optimized using Adam with a learning rate of $1{\times}10^{-4}$, weight decay of $1{\times}10^{-4}$, and batch size 128 for 5 epochs across all three datasets. The choice of 5 epochs was determined empirically by monitoring both training loss and validation performance during preliminary experiments. The training loss converged within the first five epochs, while validation performance reached a stable plateau, with additional training beyond 5 epochs yielding no appreciable improvement. Thus, 5 epochs were sufficient to achieve convergence without unnecessary additional computation. The \ONMS{} is trained separately with Adam at a learning rate of $5{\times}10^{-4}$, weight decay of $1{\times}10^{-4}$, and batch size 128. Loss coefficients are $\alpha{=}\beta{=}1$, $w_d{=}1.0$, and $w_{kl}{=}10^{-4}$. The anchor-matching IoU threshold is $\theta_m{=}0.5$, the \ONMS{} suppression threshold is $\theta_s{=}0.1$, and the overlap threshold is $\theta_o{=}0.3$. All experiments are conducted on a single NVIDIA RTX 5060 Ti, 16GB GPU.

\subsection{Results and Analysis}
\label{sec:results and analysis}

\noindent\textbf{Comparison with Existing Approaches:}
\label{sec:comparison}
We compare against three representative OnTAL frameworks, \OAT{}~\cite{kim2022oat}, \HAT{}~\cite{reza2024hat} and MATR~\cite{song2024online}. \OAT{} and \HAT{} follow the anchor-transformer paradigm and are therefore directly comparable to our approach, while MATR represents a different design paradigm. For MUSES, we additionally compare against CAG-QIL~\cite{kang2021cagqil}, the first OnTAL grouping-based baseline.

\noindent \textbf{Results on PREGO datasets:}
As shown in Table~\ref{tab:egtea_epic}, \DUGO{} achieves state-of-the-art performance on both procedural egocentric benchmarks. On EGTEA, we attain an average mAP of \textbf{24.3\%}, surpassing MATR~\cite{song2024online} by $+8.7\%$, \OAT{}~\cite{kim2022oat} by $+4.7\%$, and \HAT{}~\cite{reza2024hat} by $+2.8\%$. On EPIC-Kitchens-100, we achieve an average mAP of \textbf{16.6\%}, outperforming
MATR~\cite{song2024online} by $+3.9\%$, \OAT{}~\cite{kim2022oat} by $+2.4\%$, and \HAT{}~\cite{reza2024hat} by $+1.3\%$. The improvements are consistent across all tIoU thresholds, with the largest absolute gains occurring at the lower thresholds, indicating that \DUGO{} recovers a substantially larger pool of correctly localized proposals. These gains are most pronounced on the PREGO benchmarks where
the temporal ordering of consecutive procedural actions makes explicit history modeling particularly valuable, validating the design of the Gaussian Memory Bank and the dual-scale temporal encoder.

\begin{table}[t]
  \centering
  \caption{mAP (\%) comparison on \textbf{MUSES \cite{liu2021muses}}.
  $^{*}$Our final configuration.}
  \label{tab:muses}
  \footnotesize
  \setlength{\tabcolsep}{3pt}

  \begin{tabular*}{\columnwidth}{@{\extracolsep{\fill}}lcccccc@{}}
    \toprule
    \textbf{Model}
      & \textbf{0.3} & \textbf{0.4} & \textbf{0.5}
      & \textbf{0.6} & \textbf{0.7} & \textbf{Avg.} \\
    \midrule
    CAG-QIL~\cite{kang2021cagqil}
      & 8.5 & 6.5 & 4.2 & 2.8 & 1.9 & 4.8 \\
    OAT~\cite{kim2022oat}
      & 16.7 & 13.0 & 10.0 & 6.3 & 3.2 & 9.8 \\
    HAT~\cite{reza2024hat}
      & 19.1 & 14.7 & 10.1 & 6.3 & 3.7 & 10.8 \\
    MATR~\cite{song2024online}
      & 23.5 & 19.3 & 14.3 & 9.4 & \textbf{5.7} & 14.4 \\
    \textbf{Ours}
      & \textbf{24.7} & \textbf{20.0} & \textbf{15.5}
      & \textbf{9.8} & 4.5 & \textbf{14.9}$^{*}$ \\
    \bottomrule
  \end{tabular*}
\end{table}

\noindent\textbf{Results on the non-PREGO benchmark:}
On MUSES (Table~\ref{tab:muses}), \DUGO{} achieves an average mAP of \textbf{14.9\%},
outperforming CAG-QIL~\cite{kang2021cagqil} by $+10.1\%$, \OAT{}~\cite{kim2022oat} by $+5.1\%$,
\HAT{}~\cite{reza2024hat} by $+4.1\%$, and MATR~\cite{song2024online} by $+0.5\%$.
Although MUSES exhibits weaker procedural dependencies than EGTEA and EPIC-Kitchens-100,
since actions in TV shows and movie dramas are largely independent rather than causally sequential,
\DUGO{} still delivers a consistent improvement on this non-egocentric benchmark.
Notably, \DUGO{} surpasses all baselines at tIoU thresholds of 0.3--0.6,
achieving 24.7\%, 20.0\%, 15.5\%, and 9.8\%, respectively, while MATR retains a slight edge
at the strictest threshold of 0.7 (5.7\% vs.\ 4.5\%).
This indicates that the dual-scale encoder and stochastic memory modeling provide benefits beyond
strictly procedural settings: the local convolutional branch captures sharp boundary cues introduced
by frequent shot cuts, while the stochastic Gaussian prototypes act as a noise-robust history summary
that is less brittle than deterministic compression when the underlying action distribution is more diverse.

\subsection{Ablation Studies}
\label{sec:ablation}

We conduct ablation studies on PREGO dataset to validate each design choice.

\noindent\textbf{Component Analysis:}
\label{sec:ablation_components}
We progressively add each proposed component to the \OAT{}~\cite{kim2022oat} baseline. The proposed framework introduces three key components. First, the \textbf{DSE} (Dual-Scale Temporal Encoder) combines a depthwise 1D convolution branch with a Transformer encoder branch, which are fused through a learned projection to jointly capture both local motion dynamics and global temporal context (Section~\ref{sec:dse}). Second, the model employs \textbf{DIoU}, a 1D Distance-IoU regression loss that penalizes not only overlap deficiencies but also center misalignment between predicted and ground-truth temporal segments, thereby improving localization precision (Section~\ref{sec:pred}). Finally, the framework incorporates the \textbf{GMB} (Gaussian Memory Bank), a stochastic Gaussian prototype bank that is dynamically adapted from encoder context and injected at two stages of the network to provide procedural history context and enhance temporal reasoning (Section~\ref{sec:GMB}).


\begin{table}[t]
  \centering
  \caption{Component analysis on EGTEA dataset.
  DSE = Dual-Scale Temporal Encoder,
  GMB = Gaussian Memory Bank. $^{*}$Our final configuration.}
  \label{tab:ablation_components}
  \footnotesize
  \setlength{\tabcolsep}{3pt}

  \begin{tabular*}{\columnwidth}{@{\extracolsep{\fill}}ccc ccccc c@{}}
    \toprule
    \textbf{DSE} & \textbf{DIoU} & \textbf{GMB}
      & \textbf{0.1} & \textbf{0.2} & \textbf{0.3}
      & \textbf{0.4} & \textbf{0.5} & \textbf{Avg.} \\
    \midrule
    -- & -- & --
      & 24.9 & 23.1 & 20.6 & 17.1 & 12.2 & 19.6 \\
    -- & -- & \cmark
      & 29.2 & 27.1 & 24.0 & 18.1 & 12.7 & 22.2 \\
    \cmark & \cmark & --
      & 29.1 & 27.2 & 24.5 & 21.0 & 15.1 & 23.4 \\
    \cmark & -- & \cmark
      & 28.5 & 26.6 & 23.9 & 20.2 & 14.5 & 22.7 \\
    \cmark & \cmark & \cmark
      & \textbf{30.9} & \textbf{28.8} & \textbf{25.6}
      & \textbf{21.1} & \textbf{14.9} & \textbf{24.3}$^{*}$ \\
    \bottomrule
  \end{tabular*}
\end{table}

\noindent Table~\ref{tab:ablation_components} confirms that each component contributes positively. \DSE{} combined with the DIoU regression term improves over the \OAT{}~\cite{kim2022oat} baseline by $+3.8\%$ average mAP by capturing both local motion dynamics and long-range temporal dependencies alongside sharper boundary localization. Adding \GMB{} on top of \DSE{}+DIoU yields a complementary $+0.9\%$ gain by conditioning anchor features on learned procedural history prototypes. All three components combined yield the best result of $24.3\%$ ($+4.7\%$ over baseline), demonstrating their complementary nature.

\noindent \textbf{Loss Function Analysis:}
We compare our loss configuration against alternatives for both classification and regression. For classification, we evaluate Standard Cross-Entropy, Standard Focal Loss~\cite{lin2017focal}, Adaptive Focal Loss~\cite{reza2024hat}, and the background-suppression focal variant (BG-SUP Focal) from GateHUB~\cite{chen2022gatehub}. For regression, we compare standard L1 against L1 augmented with our 1D DIoU term. Our final configuration, Standard Focal Loss ($\gamma{=}2$) combined with L1 + 1D DIoU regression, achieves the best performance across all tIoU thresholds (Table~\ref{tab:ablation_loss}). Adding the 1D DIoU term consistently improves the overall performance across all classification losses. For Cross-Entropy, however, the mAP shows a slight decrease at tIoU~0.3 and 0.5, although the average performance still improves. L1 loss independently supervises the two normalized regression offsets, while DIoU is applied to the decoded temporal segment and jointly accounts for the IoU gap and center misalignment. The two losses therefore complement each other: L1 provides precise gradients for each offset, whereas DIoU helps address geometric errors that may remain even when the individual offset errors are small. Among the classification losses, Standard Focal Loss performs better than Cross-Entropy by reducing the influence of easy background samples. This is particularly useful for egocentric datasets, where foreground and background samples are highly imbalanced. It also achieves better results than the more aggressive BG-SUP and Adaptive Focal variants.

     

\begin{table}[t]
  \centering
  \caption{Loss function analysis on EGTEA dataset.
  $^{*}$Our final configuration.}
  \label{tab:ablation_loss}
  \footnotesize
  \setlength{\tabcolsep}{3pt}

  \begin{tabular*}{\columnwidth}{@{\extracolsep{\fill}}ll ccccc c@{}}
    \toprule
    \textbf{Cls. Loss} & \textbf{Reg. Loss}
      & \textbf{0.1} & \textbf{0.2} & \textbf{0.3}
      & \textbf{0.4} & \textbf{0.5} & \textbf{Avg.} \\
    \midrule
    \multirow{2}{*}{BG-SUP Focal}
      & L1
      & 30.2 & 27.9 & 24.3 & 19.9 & 13.8 & 23.2 \\
      & L1 + DIoU
      & 30.8 & 28.7 & 25.3 & 20.3 & 14.8 & 24.0 \\
    \midrule
    \multirow{2}{*}{Cross-Entropy}
      & L1
      & 28.7 & 27.2 & 24.1 & 19.6 & 14.4 & 22.8 \\
      & L1 + DIoU
      & 29.7 & 27.8 & 24.0 & 19.7 & 14.3 & 23.1 \\
    \midrule
    \multirow{2}{*}{Adaptive Focal}
      & L1
      & 28.9 & 26.8 & 23.2 & 19.5 & 13.7 & 22.4 \\
      & L1 + DIoU
      & 30.0 & 27.9 & 24.7 & 20.9 & 14.2 & 23.5 \\
    \midrule
    \multirow{2}{*}{Standard Focal}
      & L1
      & 28.5 & 26.6 & 23.9 & 20.2 & 14.5 & 22.7 \\
      & \textbf{L1 + DIoU}
      & \textbf{30.9} & \textbf{28.8} & \textbf{25.6}
      & \textbf{21.1} & \textbf{14.9} & \textbf{24.3}$^{*}$ \\
    \bottomrule
  \end{tabular*}
\end{table}

\noindent \textbf{Inference Speed and Early Detection:} We further evaluate the efficiency and early response capability of online temporal action localization models using inference speed (FPS), Average Early Detected Time (AEDT)~\cite{kim2022oat}, parameter count, and computational cost (GFLOPs). AEDT measures how many seconds before the end of an action a correct prediction is generated, where more negative values indicate earlier detection and are therefore preferable. To ensure a fair comparison, all methods are evaluated under the same GPU configuration. \OAT{}~\cite{kim2022oat} achieves the highest throughput at 85.27 FPS and the earliest detection time of $-1.28$,s, with 92.43M parameters and 5.545 GFLOPs. \HAT{}~\cite{reza2024hat} operates at 42.7 FPS with an AEDT of $-1.14$,s, using 247.10M parameters and 6.435 GFLOPs, while \DUGO{} reaches 40.2 FPS and an AEDT of $-1.02$,s with 129.22M parameters and 7.755 GFLOPs. Although \DUGO{} incurs a slightly higher computational cost than \OAT{} and \HAT{} in terms of GFLOPs and has marginally lower throughput, it substantially reduces the parameter count compared with \HAT{} while maintaining real-time processing speed and consistently detecting actions more than one second before completion. Overall, these results demonstrate that \DUGO{} provides a favorable balance between model size, computational cost, inference efficiency, and early action anticipation, making it well suited for practical online deployment scenarios.

\begin{figure}[t]
\centering
\includegraphics[width=\columnwidth]{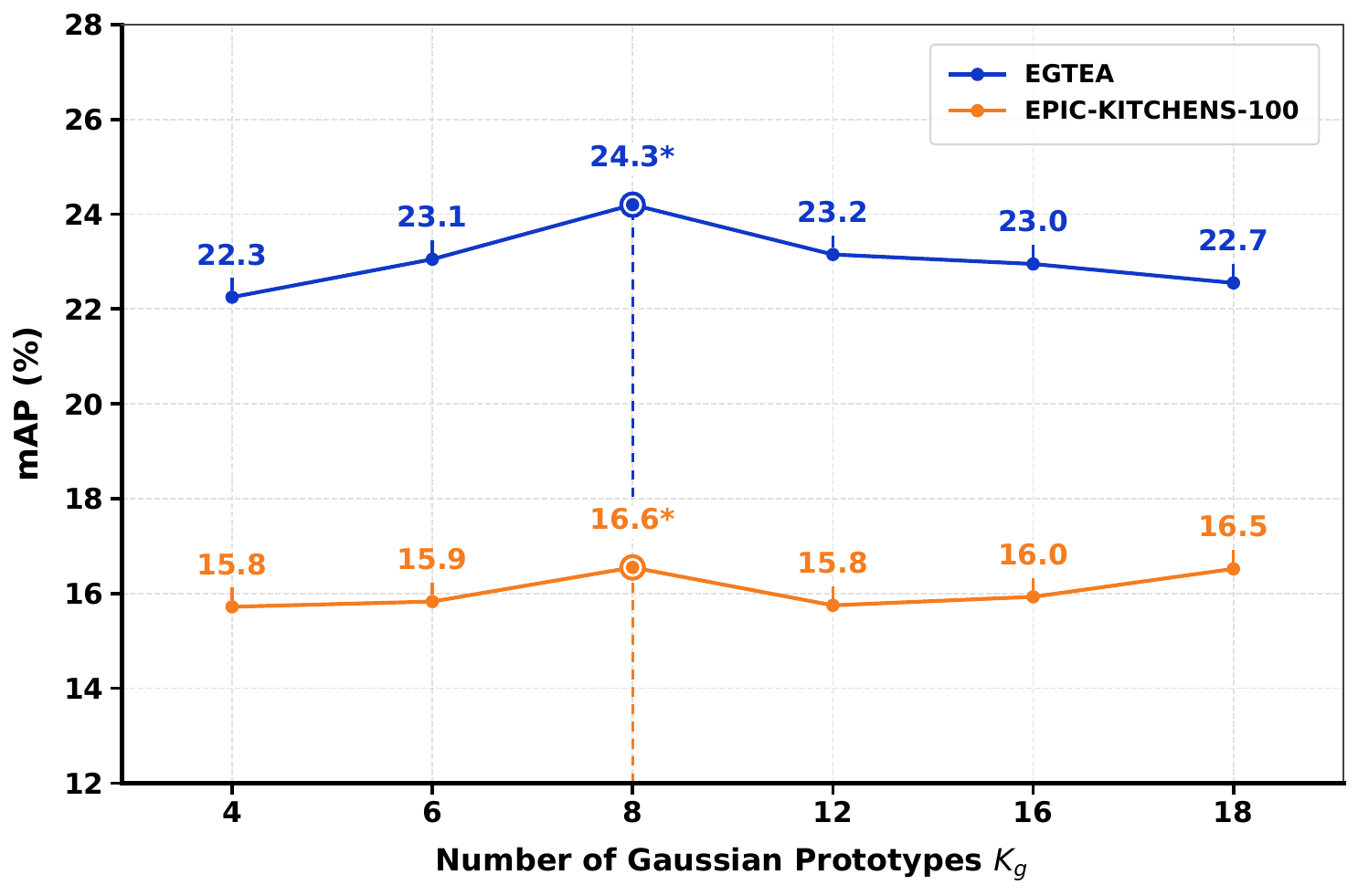}
\caption{Gaussian prototypes $K_g$ analysis of \DUGO{} on EGTEA and EPIC-Kitchens-100 datasets across various tIoU thresholds.}
\label{fig:kg_ablation}
\end{figure}

\noindent \textbf{Effect of Gaussian Prototype Count:}
Figure~\ref{fig:kg_ablation} reports the impact of varying the number of Gaussian prototypes $K_g$ in the \GMB{} on EGTEA and EPIC-Kitchens-100 across multiple tIoU thresholds. Performance consistently improves as $K_g$ increases from a single prototype to a moderate count, indicating
that multiple prototypes better capture the diverse procedural hand-motion patterns and temporal interaction dynamics present in egocentric actions. However, the gains gradually saturate once the dominant feature modes are sufficiently covered, making additional prototypes increasingly redundant. We also observe a slight performance drop when using more than eight prototypes, which may stem from over-fragmentation of the feature
space and noisier prototype assignments that weaken generalization. Overall, the chosen $K_g{=}8$ corresponds to the peak performance region on both datasets while keeping the memory bank compact and parameter-efficient. 

\subsection{Discussion}
\label{sec:discussion}

\noindent \textbf{Additional Memory Baseline Comparison:}
Table~\ref{tab:main} benchmarks \GMB{} against a range of established history-modeling strategies on EGTEA, giving us a clearer picture of how different approaches to temporal context stack up.
The no-history baseline lags well behind every memory-based method, confirming that some form of temporal context is indispensable for procedural action localization rather than optional.
Adding a recurrent summarization mechanism, as in LSTR~\cite{xu2021lstr}, helps modestly, while the attention-gated memory in GateHUB~\cite{chen2022gatehub} pushes performance a bit further still.
\HAT{}'s Fixed-Size History Memory (FSHM)~\cite{reza2024hat} does better yet, which we attribute to its anchor-aware approach to selecting relevant history.
MATR~\cite{song2024online}, interestingly, falls well short of every history-aware variant in this controlled setting, reinforcing just how much a dedicated memory module matters for reasoning over procedural sequences.
\GMB{} comes out ahead of all of these, improving over the strongest baseline (FSHM) by a solid margin and widening further against GateHUB, LSTR, and the no-history setup.
Taken together, these results indicate that a stochastic, Gaussian prototype-based memory represents history in a more expressive and generalizable manner than deterministic compression or recurrent summarization, a distinction that holds even prior to the integration of the complete \DUGO{} pipeline.


\begin{table}[!t]
  \centering
  \caption{Additional baselines on EGTEA dataset, re-implemented under a controlled setting. $^{*}$Our final configuration. HAT figures here differ from Table~\ref{tab:egtea_epic} due to this controlled re-implementation.}
  \label{tab:main}
  \footnotesize
  \setlength{\tabcolsep}{3pt}

  \begin{tabular*}{\columnwidth}{@{\extracolsep{\fill}}lcccccc@{}}
    \toprule
    & 0.1 & 0.2 & 0.3 & 0.4 & 0.5 & \textbf{Avg} \\
    \midrule
    MATR~\cite{song2024online}
      & 21.5 & 19.5 & 15.9 & 12.6 & 8.6 & 15.6 \\
    No-History
      & 24.2 & 22.7 & 20.5 & 16.3 & 10.9 & 18.9 \\
    LSTR~\cite{xu2021lstr}
      & 25.3 & 23.3 & 20.4 & 16.4 & 12.0 & 19.5 \\
    GateHUB~\cite{chen2022gatehub}
      & 26.2 & 24.0 & 21.0 & 16.6 & 11.7 & 19.9 \\
    FSHM(HAT)~\cite{reza2024hat}
      & 27.3 & 25.3 & 21.6 & 16.9 & \textbf{12.8} & 20.8 \\
    \textbf{GMB}
      & \textbf{29.2} & \textbf{27.1} & \textbf{24.0}
      & \textbf{18.1} & 12.7 & \textbf{22.2}$^{*}$ \\
    \bottomrule
  \end{tabular*}
\end{table}

\noindent \textbf{Overall Discussion:}
\DUGO{} consistently achieves state-of-the-art performance across all three benchmarks, demonstrating that its three contributions are individually meaningful and mutually complementary. The improvements are particularly pronounced on the PREGO datasets, where the temporal ordering of consecutive procedural actions makes explicit history modeling especially valuable. \DUGO{} also performs strongly on the non-procedural MUSES benchmark, showing that the proposed dual-scale encoder and stochastic memory bank generalize beyond egocentric procedural settings.

The ablation studies further show that \DSE{} and \GMB{} capture complementary information. \DSE{} strengthens the per-timestep representation by jointly modeling local motion and global context, while \GMB{} enriches anchor queries with cross-temporal procedural prototypes. Their combination therefore provides complementary benefits rather than simply duplicating the same information. The 1D DIoU loss further improves boundary precision by accounting for center misalignment, which standard L1 regression may not adequately capture when offset magnitudes are small.

\section{Conclusion}
\label{sec:conclusion}

We presented \DUGO{}, an online temporal action localization framework that integrates three complementary contributions within a unified architecture: a Dual-Scale Temporal Encoder (\DSE{}) combining depthwise temporal convolutions and transformer attention to capture both local and global temporal dependencies, a Gaussian Memory Bank (\GMB{}) that injects stochastic Gaussian prototypes into both encoder memory and anchor queries to enrich procedural context, and a 1D DIoU regression loss for improved temporal boundary localization.
A current limitation of \DUGO{} is that \GMB{} adapts prototypes only from the current short-term encoder window, limiting persistent long-horizon procedural reasoning in repetitive activities. Future work will explore recurrent step-aware memory and object-centric representations to improve long-range procedural understanding while preserving online efficiency.
Extensive experiments on EGTEA, EPIC-Kitchens-100, and MUSES demonstrate that \DUGO{} consistently outperforms prior state-of-the-art methods while maintaining strong generalization across procedural and non-procedural settings. Ablation studies further verify that each component contributes independently, with \GMB{} providing the largest gains on procedural egocentric benchmarks where explicit history modeling is particularly beneficial.


\noindent\textbf{Acknowledgement:} This work was supported by the Conference Travel and Research Grants (CTRG) 2025–2026 from North South University, under the Grant ID: CTRG-25-SEPS-31.

\bibliographystyle{IEEEtran}
\bibliography{refs}

\end{document}